\documentclass{article}

\PassOptionsToPackage{numbers, sort&compress}{natbib}

\usepackage[final]{nips_2017} % final version

\usepackage[utf8]{inputenc} % allow utf-8 input
\usepackage[T1]{fontenc}    % use 8-bit T1 fonts
\usepackage[pagebackref]{hyperref}       % hyperlinks
\usepackage{url}            % simple URL typesetting
\usepackage{booktabs}       % professional-quality tables
\usepackage{amsfonts}       % blackboard math symbols
\usepackage{nicefrac}       % compact symbols for 1/2, etc.
\usepackage{microtype}      % microtypography

\usepackage{epsfig,subfigure,graphicx,wrapfig}
\usepackage{amsmath,amssymb}
\usepackage{array,booktabs,bm,enumitem}
\usepackage{mathtools}
\usepackage{algorithm,algorithmic}
\usepackage{authblk}

\let\oldbibliography\thebibliography
\renewcommand{\thebibliography}[1]{\oldbibliography{#1}
\setlength{\itemsep}{3.4pt}}

\DeclarePairedDelimiter{\ceil}{\lceil}{\rceil}

\newcommand{\op}{\operatorname}
\newcommand{\lt}{\left}
\newcommand{\rt}{\right}

\newcommand{\PreserveBackslash}[1]{\let\temp=\\#1\let\\=\temp}
\newcolumntype{C}[1]{>{\PreserveBackslash\centering}p{#1}}
\newcolumntype{R}[1]{>{\PreserveBackslash\raggedleft}p{#1}}
\newcolumntype{L}[1]{>{\PreserveBackslash\raggedright}p{#1}}

\makeatletter
\renewcommand\AB@affilsepx{, \protect\Affilfont}
\makeatother

\usepackage{color}

\title{Wider and Deeper, Cheaper and Faster:\\ Tensorized LSTMs for Sequence Learning}

\author[1,2]{\textbf{Zhen He}}
\author[3]{\textbf{Shaobing Gao}}
\author[2]{\textbf{Liang Xiao}}
\author[2]{\textbf{Daxue Liu}}
\author[2]{\textbf{Hangen He}}
\author[1,4]{\textbf{David Barber}\hspace{9pt}\thanks{Corresponding authors: Shaobing Gao <gaoshaobing@scu.edu.cn> and Zhen He <hezhen.cs@gmail.com>.}\hspace{-9pt}}

\affil[1]{University College London}
\affil[2]{National University of Defense Technology}
\affil[3]{Sichuan University}
\affil[4]{Alan Turing Institute}

\begin{document}

\maketitle
\setcounter{footnote}{0}

\begin{abstract}
% LSTM, sequence learning
Long Short-Term Memory (LSTM) is a popular approach to boosting the ability of Recurrent Neural Networks to store longer term temporal information.
% how to enhance LSTM
The \emph{capacity} of an LSTM network can be increased by widening and adding layers.
% difficulties
However, usually the former introduces additional parameters, while the latter increases the runtime.
% our methods
As an alternative we propose the \emph{Tensorized LSTM} in which the hidden states are represented by \emph{tensors} and updated via a \emph{cross-layer convolution}.
% our advantages
By increasing the tensor size, the network can be widened efficiently without additional parameters since the parameters are shared across different locations in the tensor;
by delaying the output, the network can be deepened implicitly with little additional runtime since deep computations for each timestep are merged into temporal computations of the sequence.
% experiments
Experiments conducted on five challenging sequence learning tasks show the potential of \mbox{the proposed model.}
\end{abstract}

\section{Introduction}

% sequence learning
We consider the time-series prediction task of producing a desired output $\bm{y}_t$ at each timestep \mbox{$t\!\in\!\{1,\ldots,T\}$} given an observed input sequence $\bm{x}_{1:t}\! =\! \{\bm{x}_{1}, \bm{x}_{2}, \cdots, \bm{x}_{t}\}$, where $\bm{x}_{t} \!\in\! \mathbb{R}^{R}$ and $\bm{y}_t \!\in \!\mathbb{R}^{S}$ are vectors\footnote{ Vectors are assumed to be in row form throughout this paper.}.
% RNN
The Recurrent Neural Network (RNN) \citep{rumelhart1986learning,elman1990finding} is a powerful model that learns how to use a hidden state vector $\bm{h}_{t} \!\in \!\mathbb{R}^{M}$ to encapsulate the relevant features of the entire input history $\bm{x}_{1:t}$ up to timestep $t$.
Let $\bm{h}^{cat}_{t-1}  \!\in\! \mathbb{R}^{R+M}$ be the concatenation of the current input $\bm{x}_{t}$ and the previous hidden state $\bm{h}_{t-1}$:
\begin{align}
& \bm{h}^{cat}_{t-1} = [\bm{x}_{t}, \bm{h}_{t-1} ]
\end{align}
The update of the hidden state $\bm{h}_{t}$ is defined as:
\begin{align}
& \bm{a}_{t} = \bm{h}^{cat}_{t-1} \bm{W}^{h} + \bm{b}^{h} \label{eq_rnn_linear}\\
& \bm{h}_{t} = \phi (\bm{a}_{t})
\end{align}
where $\bm{W}^{h}\! \in\! \mathbb{R}^{(R+M) \!\times \!M}$ is the weight, $\bm{b}^{h}\! \in\! \mathbb{R}^{M}$ the bias, $\bm{a}_t\! \in\! \mathbb{R}^{M}$ the hidden activation, and $\phi(\cdot)$ the element-wise tanh function. Finally, the output $\bm{y}_{t}$ at timestep $t$ is generated by:
\begin{equation}
\bm{y}_{t} = \varphi(\bm{h}_{t} \bm{W}^{y} + \bm{b}^{y} )
\end{equation}
where $\bm{W}^{y} \!\in\! \mathbb{R}^{M \!\times \!S}$ and $\bm{b}^{y} \in \mathbb{R}^{S}$, and $\varphi(\cdot)$ can be any differentiable function, depending on the task.

% LSTM
However, this vanilla RNN has difficulties in modeling long-range dependencies due to the vanishing/exploding gradient problem \citep{bengio1994learning}.
Long Short-Term Memories (LSTMs) \citep{hochreiter1997long,gers2000learning} alleviate these problems by employing memory cells to preserve information for longer, and adopting gating mechanisms to modulate the information flow.
Given the success of the LSTM in sequence modeling, it is natural to consider how to increase the complexity of the model and thereby increase the set of tasks for which the LSTM can be profitably applied.

% capacity
We consider the \emph{capacity} of a network to consist of two components: the \emph{width} (the amount of information handled in parallel) and the \emph{depth} (the number of computation steps) \citep{bengio2009learning}.
% width problem
A naive way to widen the LSTM is to increase the number of units in a hidden layer;
however, the parameter number scales quadratically with the number of units.
% depth problem
To deepen the LSTM, the popular Stacked LSTM  (\textit{s}LSTM) stacks multiple LSTM layers \citep{graves2013speech};
however, runtime is proportional to the number of layers and information from the input is potentially lost (due to gradient vanishing/explosion) as it propagates vertically through the layers.

% contribution
In this paper, we introduce a way to both widen and deepen the LSTM whilst keeping the parameter number and runtime largely unchanged. In summary, we make the following contributions:
\vspace{-6pt}
\begin{itemize}[leftmargin=16.5pt]\setlength{\itemsep}{0pt}
  \item[(a)] We tensorize RNN hidden state vectors into higher-dimensional tensors which allow more flexible parameter sharing and can be widened more efficiently without additional parameters.
  \item[(b)] Based on (a), we merge RNN deep computations into its temporal computations so that the network can be deepened with little additional runtime, resulting in a \emph{Tensorized RNN (\textit{t}RNN)}.
  \item[(c)] We extend the \textit{t}RNN to an LSTM, namely the \emph{Tensorized LSTM (\textit{t}LSTM)}, which integrates a novel memory cell convolution to help to prevent the vanishing/exploding gradients.
\end{itemize}

\section{Method}

\subsection{Tensorizing Hidden States}

% motivation: reduce parameter number
It can be seen from (\ref{eq_rnn_linear}) that in an RNN, the parameter number scales quadratically with the size of the hidden state.
% method 1: tensor factorization
A popular way to limit the parameter number when widening the network is to organize parameters as higher-dimensional tensors which can be factorized into lower-rank sub-tensors that contain significantly fewer elements \citep{taylor2009factored,sutskever2011generating,denil2013predicting,irsoy2014modeling,novikov2015tensorizing,wu2016multiplicative,bertinetto2016learning,garipov2016ultimate,krause2016multiplicative}, which is is known as tensor factorization.
This implicitly widens the network since the hidden state vectors are in fact broadcast to interact with the tensorized parameters.
% method 2: parameter sharing
Another common way to reduce the parameter number is to share a small set of parameters across different locations in the hidden state, similar to Convolutional Neural Networks (CNNs) \citep{lecun1989backpropagation,lecun1998gradient}.

% the reason of using parameter sharing
We adopt parameter sharing to cutdown the parameter number for RNNs, since compared with factorization, it has the following advantages:
(i) \emph{scalability}, i.e., the number of shared parameters can be set independent of the hidden state size, and
(ii) \emph{separability}, i.e., the information flow can be carefully managed by controlling the \emph{receptive field}, allowing one to shift RNN deep computations to the temporal domain (see Sec.~\ref{sec_merge}).
% the reason of using tensors
We also explicitly tensorize the RNN hidden state vectors, since compared with vectors, tensors have a better:
(i) \emph{flexibility}, i.e., one can specify which dimensions to share parameters and then can just increase the size of those dimensions without introducing additional parameters, and
(ii) \emph{efficiency}, i.e., with higher-dimensional tensors, the network can be widened faster w.r.t. its depth when fixing the parameter number (see Sec.~\ref{sec_lstm}).

% 2D tensors
For ease of exposition, we first consider 2D tensors (matrices): we tensorize the hidden state $\bm{h}_{t} \!\in\! \mathbb{R}^{M}$ to become $\bm{H}_{t} \!\in\! \mathbb{R}^{P \!\times \!M}$, where $P$ is the \emph{tensor size}, and $M$ the \emph{channel size}.
% how to share parameters
We locally-connect the first dimension of $\bm{H}_{t}$ in order to share parameters, and fully-connect the second dimension of $\bm{H}_{t}$ to allow global interactions.
This is analogous to the CNN which fully-connects one dimension (e.g., the RGB channel for input images) to globally fuse different feature planes.
Also, if one compares $\bm{H}_{t}$ to the hidden state of a Stacked RNN (\textit{s}RNN) (see Fig.~\ref{fig-tlstm}(a)), then $P$ is akin to the number of stacked hidden layers, and $M$ the size of each hidden layer.
% higher-dimensional tensors
We start to describe our model based on 2D tensors, and finally show how to strengthen the model with higher-dimensional tensors.

\subsection{Merging Deep Computations}\label{sec_merge}

\begin{figure}[!t]
\centering
%\belowcaptionskip=-10pt
%\abovecaptionskip=9pt
\includegraphics[width=0.994\textwidth]{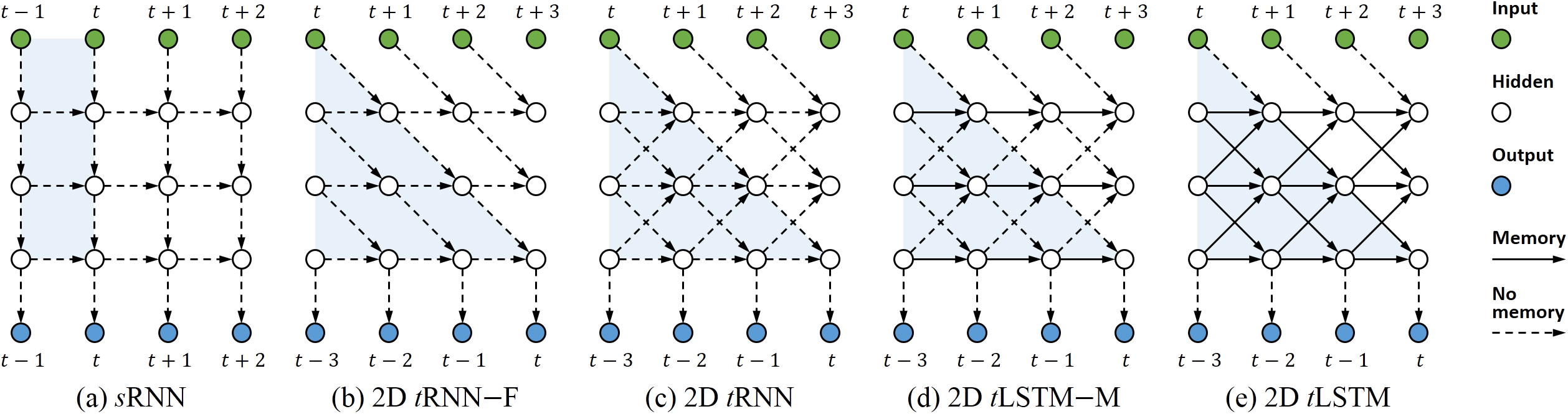}
\caption{
Examples of \textit{s}RNN, \textit{t}RNNs and \textit{t}LSTMs.
(a) A 3-layer \textit{s}RNN.
(b) A 2D \textit{t}RNN without (--) feedback (F) connections, which can be thought as a \emph{skewed} version of (a).
(c) A 2D \textit{t}RNN.
(d) A 2D \textit{t}LSTM without (--) memory (M) cell convolutions.
(e) A 2D \textit{t}LSTM.
In each model, the blank circles in column 1 to 4 denote the hidden state at timestep $t\!-\!1$ to $t\!+\!2$, respectively, and the blue region denotes the receptive field of the current output $\bm{y}_t$.
In (b)-(e), the outputs are delayed by $L\!-\!1\!=\!2$ timesteps, where $L\!=\!3$ is the depth.
}
\label{fig-tlstm}
\end{figure}

% delay for deepening
Since an RNN is already \emph{deep} in its temporal direction, we can deepen an input-to-output computation by associating the input $\bm{x}_t$ with a (delayed) future output.
% requirement
In doing this, we need to ensure that the output $\bm{y}_{t}$ is \emph{separable}, i.e., not influenced by any future input $\bm{x}_{t'}$ ($t' \!> \!t$).
% how to delay
Thus, we concatenate the projection of $\bm{x}_{t}$ to the \emph{top} of the previous hidden state $\bm{H}_{t-1}$, then gradually shift the input information down when the temporal computation proceeds, and finally generate $\bm{y}_{t}$ from the \emph{bottom} of $\bm{H}_{t+L-1}$, where $L \! - \!1$ is the number of delayed timesteps for computations of \emph{depth} $L$.
% example
An example with $L\! = \!3$ is shown in Fig.~\ref{fig-tlstm}(b). This is in fact a \emph{skewed} \textit{s}RNN as used in \citep{appleyard2016optimizing} (also similar to \citep{van2016pixel}).
% advantage over other methods
However, our method does not need to change the network structure and also allows different kinds of interactions as long as the output is separable, e.g, one can increase the local connections and use feedback (see Fig.~\ref{fig-tlstm}(c)), which can be beneficial for \textit{s}RNNs \citep{chung2015gated}.
% convolution for widening
In order to share parameters, we update $\bm{H}_{t}$ using a convolution with a learnable kernel.
% advantage of delay and convolution
In this manner we increase the complexity of the input-to-output mapping (by delaying outputs) and limit parameter growth (by sharing transition parameters using convolutions).

% concatenate input
To describe the resulting \textit{t}RNN model, let $\bm{H}^{cat}_{t-1} \!\in\! \mathbb{R}^{(P+1) \!\times \! M}$ be the concatenated hidden state, and $p \! \in \!\mathbb{Z}_+$ the \emph{location} at a tensor. The channel vector $\bm{h}^{cat}_{t-1, p} \!\in\! \mathbb{R}^M$ at location $p$ of $\bm{H}^{cat}_{t-1}$ is defined as:
\begin{equation}\label{eq_trnn_in}
\bm{h}^{cat}_{t-1,p} =
\begin{cases}
\bm{x}_t \bm{W}^{x} + \bm{b}^{x} & \text{if}~~p = 1 \\
\bm{h}_{t-1,p-1} & \text{if}~~p > 1
\end{cases}
\end{equation}
where  $\bm{W}^{x} \in \mathbb{R}^{R \!\times \!M}$ and $\bm{b}^{x} \in \mathbb{R}^{M}$.
% update tensor
Then, the update of tensor $\bm{H}_t$ is implemented via a convolution:
\begin{align}
 & \bm{A}_t = \bm{H}^{cat}_{t-1} \circledast \{\bm{W}^{h}, \bm{b}^{h} \} \label{eq_trnn_conv} \\
 & \bm{H}_t = \phi ( \bm{A}_t ) \label{eq_trnn_tanh}
\end{align}
where $\bm{W}^{h} \!\in\! \mathbb{R}^{K \!\times\! M^i \!\times \!M^o}$ is the \emph{kernel weight} of size $K$, with $M^i \!= \!M$ input channels and $M^o\! =\!M$ output channels, $\bm{b}^{h} \!\in \!\mathbb{R}^{M^o}$ is the \emph{kernel bias}, $\bm{A}_t \!\in\! \mathbb{R}^{P\! \times \! M^o}$ is the hidden activation, and $\circledast$ is the convolution operator (see Appendix~\ref{app_convh} for a more detailed definition).
Since the kernel convolves across different hidden layers, we call it the \emph{cross-layer convolution}.
The kernel enables interaction, both bottom-up and top-down across layers.
% generate output
Finally, we generate $\bm{y}_{t}$ from the channel vector $\bm{h}_{t+L-1,P}\!\in\! \mathbb{R}^M$ which is located at the \emph{bottom} of $\bm{H}_{t+L-1}$:
\begin{equation}\label{eq_trnn_out}
\bm{y}_{t} = \varphi (\bm{h}_{t+L-1,P}  \bm{W}^{y} + \bm{b}^{y} )
\end{equation}
where $\bm{W}^{y} \! \in \! \mathbb{R}^{M \!\times\! S}$ and $\bm{b}^{y}\! \in \!\mathbb{R}^{S}$.
% constraint of L, P, and K
To guarantee that the \emph{receptive field} of $\bm{y}_t$ only covers the current and previous inputs $\bm{x}_{1:t}$ (see Fig.~\ref{fig-tlstm}(c)), $L$, $P$, and $K$ should satisfy the constraint:
\begin{equation}\label{eq_delay}
 L = \ceil[\Big]{\frac{2 P}{K - K \bmod 2}}
\end{equation}
where $\ceil{\cdot}$ is the ceil operation. For the derivation of (\ref{eq_delay}), please see Appendix~\ref{app_const}.

% advantages
We call the model defined in (\ref{eq_trnn_in})-(\ref{eq_trnn_out}) the \emph{Tensorized RNN (\textit{t}RNN)}.
% parameter efficiency
The model can be widened by increasing the tensor size $P$, whilst the parameter number remains fixed (thanks to the convolution).
% runtime efficiency
Also, unlike the \textit{s}RNN of runtime complexity $O(TL)$, \textit{t}RNN breaks down the runtime complexity to $O(T\!+\!L)$, which means either increasing the sequence length $T$ or the network depth $L$ would not significantly increase the runtime.

\subsection{Extending to LSTMs}\label{sec_lstm}

% motivation
To allow the \textit{t}RNN to capture long-range temporal dependencies, one can straightforwardly extend it to an LSTM by replacing the \textit{t}RNN tensor update equations of (\ref{eq_trnn_conv})-(\ref{eq_trnn_tanh}) as follows:
\begin{align}
 & [\bm{A}^g_t, \bm{A}^i_t, \bm{A}^f_t, \bm{A}^o_t] = \bm{H}^{cat}_{t-1} \circledast \{\bm{W}^h, \bm{b}^h \} \label{eq_lstm_conv} \\
 & [\bm{G}_t, \bm{I}_t, \bm{F}_t, \bm{O}_t ]  =  [\phi (\bm{A}^g_t), \sigma (\bm{A}^i_t),  \sigma(\bm{A}^f_t), \sigma (\bm{A}^o_t)]  \\
 & \bm{C}_t = \bm{G}_t \odot \bm{I}_t  + \bm{C}_{t-1} \odot \bm{F}_t \label{eq_lstm_mem}\\
 & \bm{H}_t = \phi (\bm{C}_t) \odot \bm{O}_t \label{eq_lstm_tanh}
\end{align}
where the kernel $\{\bm{W}^h, \bm{b}^h\}$ is of size $K$, with $M^i \!\!= \!\!M$ input channels and $M^o\!\! =\!4M$ output channels, $\bm{A}^g_t,\! \bm{A}^i_t,\! \bm{A}^f_t,\! \bm{A}^o_t\! \in \!\mathbb{R}^{P \!\times\! M}$ are activations for the new content $\bm{G}_t$, input gate $\bm{I}_t$, forget gate $\bm{F}_t$, and output gate $\bm{O}_t$, respectively, $\sigma(\cdot) $ is the element-wise sigmoid function, and $\bm{C}_t\! \in \!\mathbb{R}^{P\! \times\! M}$ is the memory cell.
% problem
However, since in (\ref{eq_lstm_mem}) the previous memory cell $\bm{C}_{t-1}$ is only gated along the temporal direction (see Fig.~\ref{fig-tlstm}(d)), long-range dependencies from the input to output might be lost when the tensor size $P$ becomes large.

%  our solution
\textbf{Memory Cell Convolution.}~~~
To capture long-range dependencies from multiple directions, we additionally introduce a novel \emph{memory cell convolution}, by which the memory cells can have a larger receptive field (see Fig.~\ref{fig-tlstm}(e)).
We also dynamically generate this convolution kernel so that it is both time- and location-dependent, allowing for flexible control over long-range dependencies from different directions.
This results in our \textit{t}LSTM tensor update equations:
\begin{align}
 & [\bm{A}^g_t, \bm{A}^i_t, \bm{A}^f_t, \bm{A}^o_t, \bm{A}^q_t] = \bm{H}^{cat}_{t-1} \circledast \{\bm{W}^h, \bm{b}^h \} \label{eq_tlstm_hid_conv}\\
 & [\bm{G}_t, \bm{I}_t, \bm{F}_t, \bm{O}_t, \bm{Q}_t ]  =  [\phi (\bm{A}^g_t), \sigma (\bm{A}^i_t), \sigma (\bm{A}^f_t), \sigma (\bm{A}^o_t), \varsigma (\bm{A}^q_t)] \label{eq_tlstm_non_linear} \\
 & \bm{W}^c_{t}(p) = \op{reshape} \lt(\bm{q}_{t,p}, \lt[K, 1, 1\rt] \rt) \label{eq_tlstm_reshape} \\
 & \bm{C}^{conv}_{t-1} = \bm{C}_{t-1} \circledast \bm{W}^c_{t}(p) \label{eq_tlstm_mem_conv} \\
 & \bm{C}_t = \bm{G}_t \odot \bm{I}_t + \bm{C}^{conv}_{t-1} \odot \bm{F}_t  \label{eq_tlstm_mem}\\
 & \bm{H}_t = \phi (\bm{C}_t) \odot \bm{O}_t \label{eq_tlstm_tanh}
\end{align}
\begin{wrapfigure}{r}{0.31\textwidth}
\centering
\belowcaptionskip=-10pt
%\abovecaptionskip=0pt
\vspace{-12pt}
\includegraphics[width=0.28\textwidth]{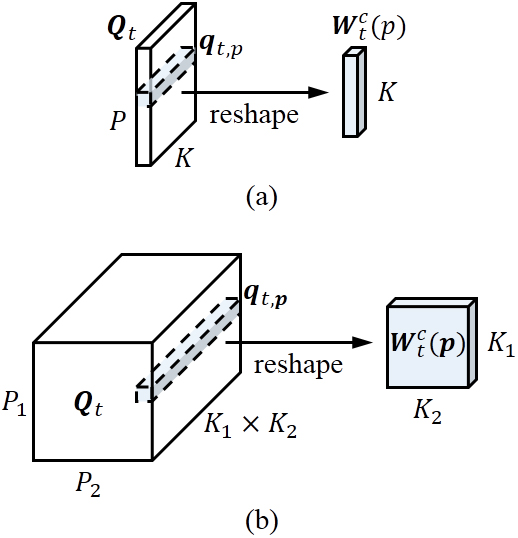}
\caption{Illustration of generating the memory cell convolution kernel, where (a) is for 2D tensors and (b) for 3D tensors.}
\label{fig-mem}
\end{wrapfigure}
where, in contrast to (\ref{eq_lstm_conv})-(\ref{eq_lstm_tanh}),
% A^q and Q
the kernel $\{\bm{W}^h, \bm{b}^h\}$ has additional $\langle K\rangle$ output channels\footnote{The operator $\langle\cdot\rangle$ returns the cumulative product of all elements in the input variable.} to generate the activation $\bm{A}^q_t\! \in \!\mathbb{R}^{P \!\times\! \langle\!K\!\rangle}$ for the dynamic kernel bank $\bm{Q}_t\! \in \!\mathbb{R}^{P \!\times\!  \langle\!K\!\rangle}$,
% q
$\bm{q}_{t,p}\!\in\!\mathbb{R}^{ \langle\!K\!\rangle}$ is the vectorized adaptive kernel at the location $p$ of $\bm{Q}_t$,
% W^c
and $\bm{W}^c_{t}(p) \!\in\! \mathbb{R}^{K \!\times 1 \times 1}$ is the dynamic kernel of size $K$ with a single input/output channel, which is reshaped from $\bm{q}_{t,p}$ (see Fig.~\ref{fig-mem}(a) for an illustration).
% memory cell convolution
In (\ref{eq_tlstm_mem_conv}), \emph{each channel} of the previous memory cell $\bm{C}_{t-1}$ is convolved with $\bm{W}^c_{t}(p)$ whose values vary with $p$, forming a \emph{memory cell convolution} (see Appendix~\ref{app_convc} for a more detailed definition), which produces a convolved memory cell $\bm{C}^{conv}_{t-1} \!\in\! \mathbb{R}^{P \!\times\! M}$.
% softmax
Note that in (\ref{eq_tlstm_non_linear}) we employ a softmax function $\varsigma (\cdot)$ to normalize the channel dimension of $\bm{Q}_t$, which, similar to \citep{leifert2016cells}, can stabilize the value of memory cells and help to prevent the vanishing/exploding gradients (see Appendix~\ref{app_grad} for details).

% comparison to similar work
The idea of dynamically generating network weights has been used in many works \citep{schmidhuber1992learning2,sutskever2011generating,denil2013predicting,bertinetto2016learning,de2016dynamic,ha2016hypernetworks}, where in \citep{de2016dynamic} location-dependent convolutional kernels are also dynamically generated to improve CNNs.
In contrast to these works, we focus on broadening the receptive field of \textit{t}LSTM memory cells. Whilst the flexibility is retained, fewer parameters are required to generate the kernel since the kernel is shared by different memory cell channels.

\textbf{Channel Normalization.}~~~
% motivation
To improve training, we adapt Layer Normalization (LN) \citep{ba2016layer} to our \textit{t}LSTM.
% why not LN
Similar to the observation in \citep{ba2016layer} that LN does not work well in CNNs where channel vectors at different locations have very different statistics, we find that LN is also unsuitable for \textit{t}LSTM where lower level information is near the input while higher level information is near the output.
% what we do
We therefore normalize the channel vectors at different locations with their own statistics, forming a \emph{Channel Normalization (CN)}, with its operator $\op{CN}\lt(\cdot\rt)$:
\begin{equation}\label{eq_cn}
\op{CN}\lt(\bm{Z}; \bm{\Gamma}, \bm{B}\rt) = \widehat{\bm{Z}} \odot \bm{\Gamma} + \bm{B}
\end{equation}
where $\bm{Z}, \widehat{\bm{Z}}, \bm{\Gamma}, \bm{B} \in \mathbb{R}^{P \times M^z}$ are the original tensor, normalized tensor, \emph{gain} parameter, and \emph{bias} parameter, respectively. The $m^z$-th channel of $\bm{Z}$, i.e. $\bm{z}_{m^z}\!\in\! \mathbb{R}^P$, is normalized element-wisely:
\begin{equation}\label{norm}
\widehat{\bm{z}}_{m^z} = (\bm{z}_{m^z} - \bm{z}^{\mu}) / \bm{z}^{\sigma} \\
\end{equation}
where $\bm{z}^{\mu}, \bm{z}^{\sigma} \!\in\! \mathbb{R}^P$ are the \emph{mean} and \emph{standard deviation} along the channel dimension of $\bm{Z}$, respectively, and $\widehat{\bm{z}}_{m^z} \!\in\! \mathbb{R}^P$ is the $m^z$-th channel of $\widehat{\bm{Z}}$.
% advantage
Note that the number of parameters introduced by CN/LN can be neglected as it is very small compared to the number of other parameters in the model.

\textbf{Using Higher-Dimensional Tensors.}~~~
% motivation
One can observe from (\ref{eq_delay}) that when fixing the kernel size $K$, the tensor size $P$ of a 2D \textit{t}LSTM grows linearly w.r.t. its depth $L$.
How can we expand the tensor volume more rapidly so that the network can be widened more efficiently?
We can achieve this goal by leveraging higher-dimensional tensors.
% method
Based on previous definitions for 2D \textit{t}LSTMs, we replace the 2D tensors with $D$-dimensional ($D\! >\! 2$) tensors, obtaining $\bm{H}_{t}, \bm{C}_{t} \!\in\! \mathbb{R}^{P_1 \! \times\!  P_2 \! \times \! \ldots\!  \times \! P_{D\!-\!1} \! \times\!  M}$ with the tensor size $ \textit{\textsf{P}} \!= \!\lt[P_1, P_2, \ldots, P_{D-1}\rt]$.
% input
Since the hidden states are no longer matrices, we concatenate the projection of $\bm{x}_t$ to one \emph{corner} of $\bm{H}_{t-1}$, and thus (\ref{eq_trnn_in}) is extended as:
\begin{equation}\label{inh}
\bm{h}^{cat}_{t-1, \bm{p}}  =
\begin{cases}
\bm{x}_t \bm{W}^x + \bm{b}^x & \text{if}~~p_d = 1~~\text{for}~~d = 1,2, \ldots, D - 1\\
\bm{h}_{t-1, \bm{p} - \bm{1}} & \text{if}~~p_d > 1~~\text{for}~~d = 1, 2, \ldots, D - 1 \\
\bm{0} & \text{otherwise}
\end{cases}
\end{equation}
where $\bm{h}^{cat}_{t-1, \bm{p}} \!\in \!\mathbb{R}^M$ is the channel vector at location $\bm{p} \!\in \! \mathbb{Z}^{D-1}_+$ of the concatenated hidden state $\bm{H}^{cat}_{t-1}\! \in \!\mathbb{R}^{(\!P_1+1\!) \!\times \!(\!P_2+1\!) \!\times\! \ldots \!\times\! (\!P_{D\!-\!1}+1\!) \!\times \!M}$.
% tensor update
For the tensor update, the convolution kernel $\bm{W}^h$ and $\bm{W}^c_t(\cdot)$ also increase their dimensionality with kernel size $\textit{\textsf{K}} =\lt[K_1, K_2, \ldots, K_{D-1}\rt]$. Note that $\bm{W}^c_t(\cdot)$ is reshaped from the vector, as illustrated in Fig.~\ref{fig-mem}(b).
% output
Correspondingly, we generate the output $\bm{y}_{t}$ from the opposite \emph{corner} of $\bm{H}_{t+L-1}$, and therefore (\ref{eq_trnn_out}) is modified as:
\begin{equation}\label{outh}
\bm{y}_{t} = \varphi (\bm{h}_{t+L-1, \textit{\textsf{P}}}  \bm{W}^{y} + \bm{b}^{y} )
\end{equation}
% depth
For convenience, we set $P_d = P$ and $K_d = K$ for $d = 1, 2, \ldots, D - 1$ so that all dimensions of \textit{\textsf{P}} and \textit{\textsf{K}} can satisfy (\ref{eq_delay}) with the same depth $L$.
% CN
In addition, CN still normalizes the channel dimension of tensors.

\section{Experiments}

% ablation
We evaluate \textit{t}LSTM on five challenging sequence learning tasks under different configurations:
\vspace{-6pt}
\begin{itemize}[leftmargin=16.5pt]\setlength{\itemsep}{0pt}
      \item[(a)] \textbf{\textit{s}LSTM (baseline)}: our implementation of \textit{s}LSTM \citep{graves2013generating} with parameters shared across all layers.
      \item[(b)] \textbf{2D \textit{t}LSTM}: the standard 2D \textit{t}LSTM, as defined in (\ref{eq_tlstm_hid_conv})-(\ref{eq_tlstm_tanh}).
      \item[(c)] \textbf{2D \textit{t}LSTM--M}: removing (--) memory (M) cell convolutions from (b), as defined in (\ref{eq_lstm_conv})-(\ref{eq_lstm_tanh}).
      \item[(d)] \textbf{2D \textit{t}LSTM--F}: removing (--) feedback (F) connections from (b).
      \item[(e)] \textbf{3D \textit{t}LSTM}: tensorizing (b) into 3D \textit{t}LSTM.
      \item[(f)] \textbf{3D \textit{t}LSTM+LN}: applying (+) LN \citep{ba2016layer} to (e).
      \item[(g)] \textbf{3D \textit{t}LSTM+CN}: applying (+) CN to (e), as defined in (\ref{eq_cn}).
\end{itemize}
\vspace{-6pt}
To compare different configurations, we also use $L$ to denote the number of layers of a \textit{s}LSTM, and $M$ to denote the hidden size of each \textit{s}LSTM layer.
We set the kernel size $K$ to 2 for 2D \textit{t}LSTM--F and 3 for other \textit{t}LSTMs, in which case we have $L \!= \!P$ according to (\ref{eq_delay}).

% parameter size invariance
For each configuration, we fix the parameter number and increase the tensor size to see if the performance of \textit{t}LSTM can be boosted without increasing the parameter number.
% runtime invariance
We also investigate how the runtime is affected by the depth, where the runtime is measured by the average GPU milliseconds spent by \emph{a forward and backward pass over one timestep of a single example}.
% sota
Next, we compare \textit{t}LSTM against the state-of-the-art methods to evaluate its ability.
% visualize
Finally, we visualize the internal working mechanism of \textit{t}LSTM.
% training details
Please see Appendix~\ref{app_train} for training details.

\pagebreak

\subsection{Wikipedia Language Modeling}

\begin{wrapfigure}{r}{0.33\textwidth}
\centering
%\belowcaptionskip=-15pt
%\abovecaptionskip=3pt
\vspace{-12pt}
\includegraphics[height=160pt]{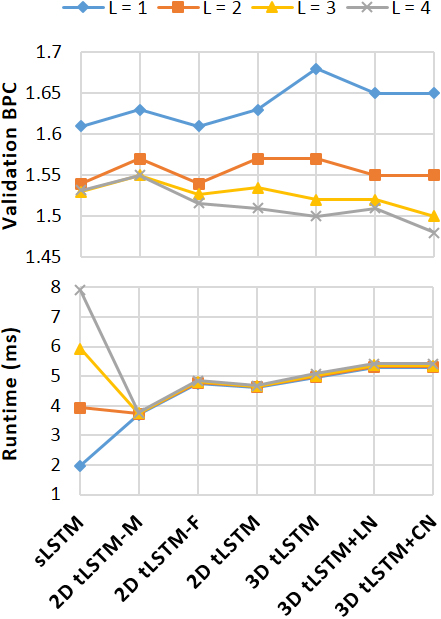}
\caption{Performance and runtime of different configurations on Wikipedia.}
\vspace{-10pt}
\label{fig-curve-wiki}
\end{wrapfigure}

% dataset
The Hutter Prize Wikipedia dataset \citep{hutterhuman} consists of 100 million characters taken from 205 different characters including alphabets, XML markups and special symbols.
We model the dataset at the character-level, and try to predict the next character of the input sequence.

% configuration
We fix the parameter number to 10M, corresponding to channel sizes $M$ of 1120 for \textit{s}LSTM and 2D \textit{t}LSTM--F, 901 for other 2D \textit{t}LSTMs, and 522 for 3D \textit{t}LSTMs.
All configurations are evaluated with depths $L\!=\! 1, 2, 3, 4$.
We use Bits-per-character (BPC) to measure the model performance.

Results are shown in Fig.~\ref{fig-curve-wiki}.
% mem
When $L \! \leq \!2$, \textit{s}LSTM and 2D \textit{t}LSTM--F outperform other models because of a larger $M$.
With $L$ increasing, the performances of \textit{s}LSTM and 2D \textit{t}LSTM--M improve but become saturated when $L \!\geq\! 3$,
while \textit{t}LSTMs with memory cell convolutions improve with increasing $L$ and finally outperform both \textit{s}LSTM and 2D \textit{t}LSTM--M.
% feedback
When $L\! = \!4$, 2D \textit{t}LSTM--F is surpassed by 2D \textit{t}LSTM,
% tensor
which is in turn surpassed by 3D \textit{t}LSTM.
% CN
The performance of 3D \textit{t}LSTM+LN benefits from LN only when $L \!\leq \!2$.
However, 3D \textit{t}LSTM+CN consistently improves 3D \textit{t}LSTM with different $L$.

\begin{wraptable}{r}{84mm}
\footnotesize
\centering
%\belowcaptionskip=0pt
%\abovecaptionskip=0pt
\vspace{-19pt}
\caption{Test BPC on Wikipedia.}
\begin{tabular*}{76mm}{L{45mm} L{8mm} C{12mm}}
\toprule
                                                                                                       & BPC    & \# Param.                                  \\
\midrule
%Stacked LSTM           \citep{graves2013generating}               &1.67  &   $\approx$27M                       \\
%GF-RNN                      \citep{chung2015gated}                         &1.58   &   $\approx$20M                       \\
%Grid LSTM                  \citep{kalchbrenner2015grid}               &1.47   &    16.8M                                      \\
MI-LSTM                     \citep{wu2016multiplicative}                 &1.44  &       $\approx$17M                        \\
mLSTM                          \citep{krause2016multiplicative}          &1.42   &    $\approx$20M                        \\
HyperLSTM+LN               \citep{ha2016hypernetworks}             &1.34   &    26.5M                       \\
HM-LSTM+LN               \citep{chung2016hierarchical}            &1.32  &      $\approx$35M                        \\
%RHN - Rec. depth 5     \citep{zilly2016recurrent}                        &1.31  &     $\approx$23M                        \\
%RHN - Rec. depth 10     \citep{zilly2016recurrent}                        &1.30  &     $\approx$21M                        \\
Large RHN                 \citep{zilly2016recurrent}                   &1.27                   &      $\approx$46M                        \\
Large FS-LSTM-4                       \citep{mujika2017fast}      &1.245                 &     $\approx$47M                        \\
2 $\times$ Large FS-LSTM-4    \citep{mujika2017fast}    &\textbf{1.198}  &    $\approx$94M                        \\
\midrule
3D \textit{t}LSTM+CN ($L\! =\! 6$, $M\! = \!1200$)  & 1.264   & 50.1M                     \\
\bottomrule
~\\[-20pt]
\end{tabular*}\label{tab_wiki}
\end{wraptable}
% runtime
Whilst the runtime of \textit{s}LSTM is almost proportional to $L$, it is nearly constant in each \textit{t}LSTM configuration and largely independent of $L$.

% sota
We compare a larger model, i.e. a 3D \textit{t}LSTM+CN with $L \! = \!6$ and $M \!= \!1200$, to the state-of-the-art methods on the test set, as reported in Table \ref{tab_wiki}.
Our model achieves 1.264 BPC with 50.1M parameters, and is competitive to the best performing methods \citep{zilly2016recurrent,mujika2017fast} with similar parameter numbers.\vspace{2.5pt}

\subsection{Algorithmic Tasks}\vspace{2.5pt}

\begin{wrapfigure}{r}{0.62\textwidth}
\centering
%\belowcaptionskip=-15pt
%\abovecaptionskip=3pt
\vspace{-12pt}
\includegraphics[height=160pt]{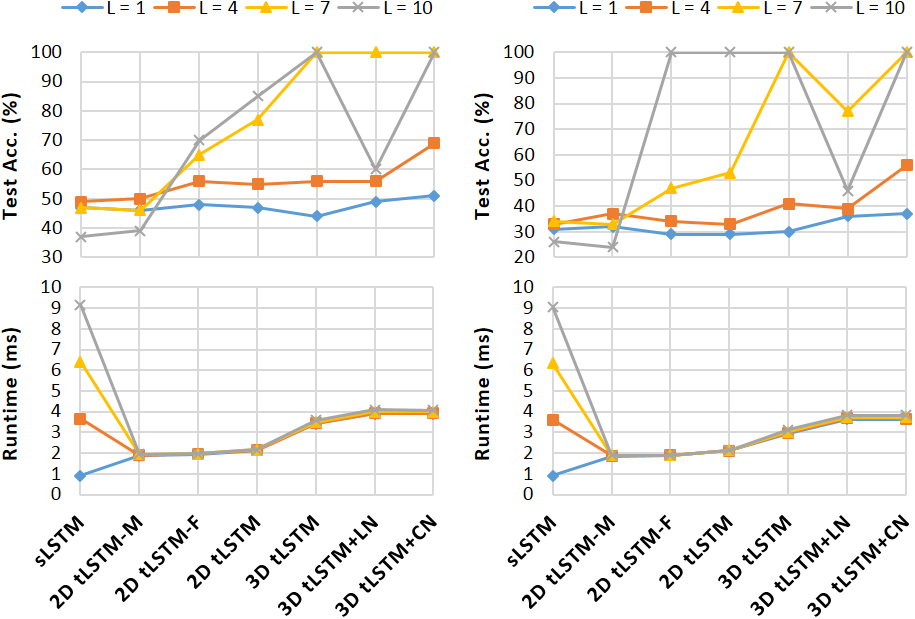}
\caption{Performance and runtime of different configurations on the addition (left) and memorization (right) tasks.}
\vspace{-20pt}
\label{fig-curve-algo}
\end{wrapfigure}

% dataset
(a) \textbf{Addition}:
The task is to sum two 15-digit integers.
The network first reads two integers with one digit per timestep, and then predicts the summation.
We follow the processing of \citep{kalchbrenner2015grid}, where a symbol `\verb"-"' is used to delimit the integers as well as pad the input/target sequence. A 3-digit integer addition task is of the form:
\begin{equation*}
\begin{aligned}
&\verb"Input:" \!\!\!\! & \verb"-"~\verb"1"~\verb"2"~\verb"3"~\verb"-"~\verb"9"~\verb"0"~\verb"0"~\verb"-"~\verb"-"~\verb"-"~\verb"-"~\verb"-" \\
&\verb"Target:" \!\!\!\! & \verb"-"~\verb"-"~\verb"-"~\verb"-"~\verb"-"~\verb"-"~\verb"-"~\verb"-"~\verb"1"~\verb"0"~\verb"2"~\verb"3"~\verb"-"
\end{aligned}
\end{equation*}

(b) \textbf{Memorization}:
The goal of this task is to memorize a sequence of 20 random symbols.
Similar to the addition task, we use 65 different symbols. A 5-symbol memorization task is of the form:
\begin{equation*}
\begin{aligned}
&\verb"Input:" & \verb"-"~\verb"a"~\verb"b"~\verb"c"~\verb"c"~\verb"b"~\verb"-"~\verb"-"~\verb"-"~\verb"-"~\verb"-"~\verb"-" \\
&\verb"Target:" & \verb"-"~\verb"-"~\verb"-"~\verb"-"~\verb"-"~\verb"-"~\verb"a"~\verb"b"~\verb"c"~\verb"c"~\verb"b"~\verb"-"
\end{aligned}
\end{equation*}

% configuration
We evaluate all configurations with $L \! = \! 1, 4, 7, 10$ on both tasks, where $M$ is 400 for \emph{addition} and 100 for \emph{memorization}.
The performance is measured by the symbol prediction accuracy.

Fig.~{\ref{fig-curve-algo}} show the results.
% mem
In both tasks, large $L$ degrades the performances of \textit{s}LSTM and 2D \textit{t}LSTM--M.
In contrast, the performance of 2D \textit{t}LSTM--F steadily improves with $L$ increasing,
% feedback
and is further enhanced by using feedback connections,
% tensor
higher-dimensional tensors,
% CN
and CN, while LN helps only when $L\!=\! 1$.
% result
Note that in both tasks, the correct  solution can be found (when $100\%$ test accuracy is achieved) due to the repetitive nature of the task.
In our experiment, we also observe that for the addition task, 3D \textit{t}LSTM+CN with $L \! = \! 7$ outperforms other configurations and finds the solution with only 298K training samples, while for the memorization task, 3D \textit{t}LSTM+CN with $L \! = \! 10$ beats others configurations and achieves perfect memorization after seeing 54K training samples.
% runtime
Also, unlike in \textit{s}LSTM, the runtime of all \textit{t}LSTMs is largely unaffected by $L$.

\begin{wraptable}{r}{94mm}
\footnotesize
\centering
%\belowcaptionskip=0pt
%\abovecaptionskip=-5pt
\vspace{-19pt}
\caption{Test accuracies on two algorithmic tasks.}
\begin{tabular*}{86mm}{L{32mm} C{6mm} C{11mm} C{6mm} C{11mm} }
\toprule
& \multicolumn{2}{c}{Addition} & \multicolumn{2}{c}{Memorization}  \\
\cmidrule(l{1.5mm}r{1.5mm}){2-3} \cmidrule(l{1.5mm}r{1.5mm}){4-5}
& Acc.     & \# Samp.  & Acc.     & \# Samp.  \\
\midrule
Stacked LSTM  \citep{graves2013generating}   &~~~~51\%   & ~~~~5M      &$>$50\%    & 900K          \\
Grid LSTM   \citep{kalchbrenner2015grid}     &$>$99\%   & 550K     &$>$99\%    & 150K     \\
\midrule
3D \textit{t}LSTM+CN ($L\! = \! 7$)       &$>$99\%   &  \textbf{298K}   &$>$99\%   & 115K        \\
3D \textit{t}LSTM+CN ($L\!  = \! 10$)       &$>$99\%   & 317K                    &$>$99\%   & ~~\textbf{54K}      \\
\bottomrule
~\\[-30pt]
\end{tabular*}
\label{tab_algo}
\end{wraptable}

% sota
\vspace{6pt}
We further compare the best performing configurations to the state-of-the-art methods for both tasks (see Table \ref{tab_algo}).
Our models solve both tasks significantly faster (i.e., using fewer training samples) than other models, achieving the new state-of-the-art results.

\subsection{MNIST Image Classification}

\begin{wrapfigure}{r}{0.62\textwidth}
\centering
%\belowcaptionskip=-15pt
%\abovecaptionskip=3pt
\vspace{-12pt}
\includegraphics[height=160pt]{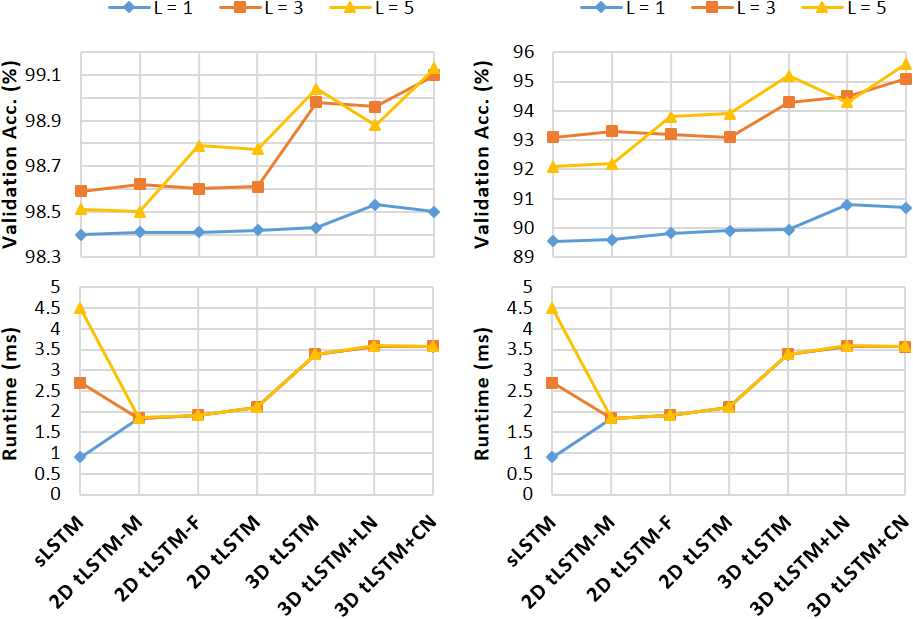}
\caption{Performance and runtime of different configurations on sequential MNIST (left) and sequential \textit{p}MNIST (right).}
\vspace{-10pt}
\label{fig-curve-mnist}
\end{wrapfigure}

% dataset
The MNIST dataset \citep{lecun1998gradient} consists of 50000/10000/10000 handwritten digit images of size $28 \!\times\! 28$ for training/validation/test. We have two tasks on this dataset:

(a) \textbf{Sequential MNIST}:
The goal is to classify the digit after sequentially reading the pixels in a scanline order \citep{le2015simple}.
It is therefore a 784 timestep sequence learning task where a single output is produced at the last timestep; the task requires very long range dependencies in the sequence.

(b) \textbf{Sequential Permuted MNIST}:
We permute the original image pixels in a fixed random order as in \citep{arjovsky2016unitary}, resulting in a permuted MNIST (\textit{p}MNIST) problem that has even longer range dependencies across pixels and is harder.

% configuration
In both tasks, all configurations are evaluated with $M\!=\!100$ and $L \! = \! 1, 3, 5$.
The model performance is measured by the classification accuracy.

Results are shown in Fig.~\ref{fig-curve-mnist}.
% mem
\textit{s}LSTM and 2D \textit{t}LSTM--M no longer benefit from the increased depth when $L \!= \!5$.
Both increasing the depth
% tensor
and tensorization boost the performance of 2D \textit{t}LSTM.
% feedback
However, removing feedback connections from 2D \textit{t}LSTM seems not to affect the performance.
% CN
On the other hand, CN enhances the 3D \textit{t}LSTM and when $L\!\geq\! 3$ it outperforms LN.
% result
3D \textit{t}LSTM+CN with $L \! =\!5$ achieves the highest performances in both tasks, with a validation accuracy of 99.1\% for MNIST and 95.6\% for \textit{p}MNIST.
% runtime
The runtime of \textit{t}LSTMs is negligibly affected by $L$, and all \textit{t}LSTMs become faster than \textit{s}LSTM when $L \!= \!5$.

\begin{wraptable}{r}{85mm}
\footnotesize
\centering
%\belowcaptionskip=-12pt
%\abovecaptionskip=0pt
\vspace{-7pt}
\caption{Test accuracies (\%) on sequential MNIST/\textit{p}MNIST.}
\begin{tabular*}{78mm}{L{38mm} C{14mm} C{14mm}}
\toprule
& MNIST    & \textit{p}MNIST  \\
\midrule
%TANH \citep{le2015simple}    &35.0   & 35.0                       \\
\textit{i}RNN \citep{le2015simple}      &97.0   & 82.0                       \\
LSTM \citep{arjovsky2016unitary}    &98.2  &88.0                        \\
\textit{u}RNN \citep{arjovsky2016unitary}   &95.1  & 91.4                     \\
Full-capacity \textit{u}RNN \citep{wisdom2016full}   &96.9  & 94.1                     \\
\textit{s}TANH \citep{zhang2016architectural}    &98.1  & 94.0                     \\
BN-LSTM \citep{cooijmans2016recurrent}    &99.0        & 95.4                \\
Dilated GRU \citep{chang2017dilated}    &\textbf{99.2}        & 94.6                \\
Dilated CNN \citep{oord2016wavenet} in \citep{chang2017dilated}    &98.3        & \textbf{96.7}                \\
\midrule
3D \textit{t}LSTM+CN ($L\! = \! 3$)                 &\textbf{99.2}        & 94.9        \\
3D \textit{t}LSTM+CN ($L\!  = \! 5$)                 &99.0                     &95.7        \\
\bottomrule
~\\[-25pt]
\end{tabular*}
\label{tab_mnist}
\end{wraptable}

% sota
We also compare the configurations of the highest test accuracies to the state-of-the-art methods (see \mbox{Table \ref{tab_mnist}}).
For sequential MNIST, our 3D \textit{t}LSTM+CN with $L\! =\!3$ performs as well as the state-of-the-art Dilated GRU model \citep{chang2017dilated}, with a test accuracy of 99.2\%.
For the sequential \textit{p}MNIST, our 3D \textit{t}LSTM+CN with $L\! =\!5$ has a test accuracy of 95.7\%, which is close to the state-of-the-art of 96.7\% produced by the Dilated CNN \citep{oord2016wavenet} in \citep{chang2017dilated}.\vspace{-5pt}

\subsection{Analysis}\vspace{-3pt}

\begin{figure}[!t]
\centering
\belowcaptionskip=-13pt
\abovecaptionskip=-9pt
\includegraphics[width=0.999\textwidth]{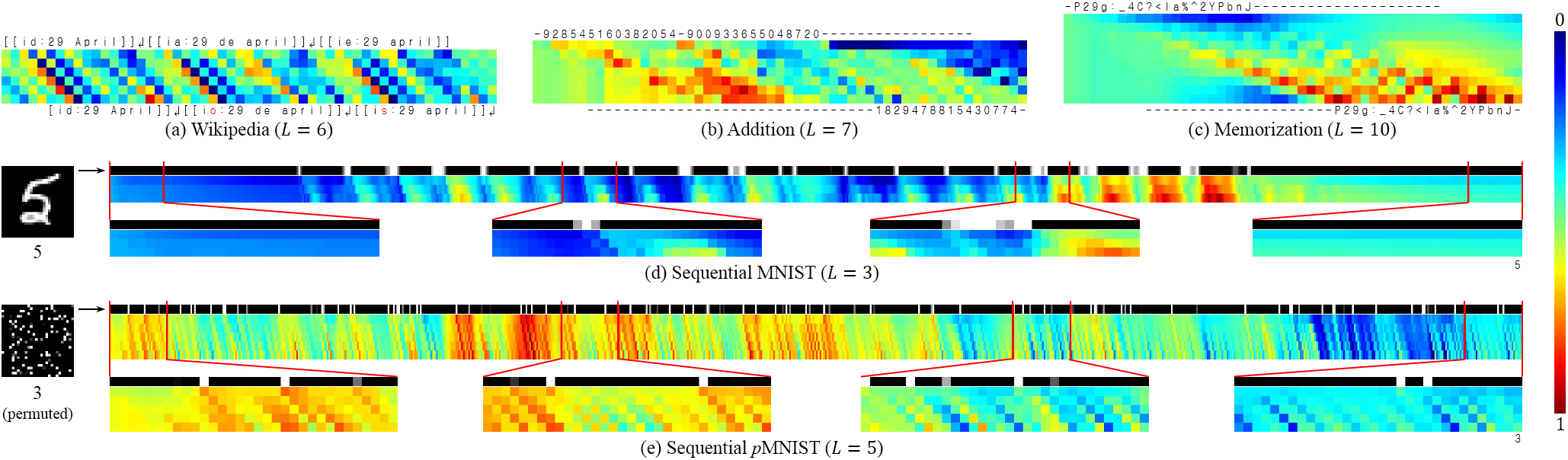}
\caption{Visualization of the diagonal channel means of the \textit{t}LSTM memory cells for each task. In each horizontal bar, the rows from top to bottom correspond to the diagonal locations from $\bm{p}^{in}$ to $\bm{p}^{out}$, the columns from left to right correspond to different timesteps (from $1$ to $T\! +\! L\! -\! 1$ for the full sequence, where $L \!- \!1$ is the time delay), and the values are normalized to be in range $[0, 1]$ for better visualization. Both full sequences in (d) and (e) are zoomed out horizontally.}
\label{fig-vis}
\end{figure}

% Advantage
% parameter, runtime
The experimental results of different model configurations on different tasks suggest that the performance of \textit{t}LSTMs can be improved by increasing the tensor size and network depth, requiring no additional parameters and little additional runtime.
% mem
As the network gets wider and deeper, we found that the memory cell convolution mechanism is crucial to maintain improvement in performance.
% feedback
Also, we found that feedback connections are useful for tasks of sequential output (e.g., our Wikipedia and algorithmic tasks).
% CN, tensor
Moreover, \textit{t}LSTM can be further strengthened via tensorization or CN.

% Visualization
It is also intriguing to examine the internal working mechanism of \textit{t}LSTM. Thus, we visualize the memory cell which gives insight into how information is routed.
For each task, the best performing \textit{t}LSTM is run on a random example.
We record the channel mean (the mean over channels, e.g., it is of size $P\! \times\! P$ for 3D \textit{t}LSTMs) of the memory cell at each timestep, and visualize the diagonal values of the channel mean from location $\bm{p}^{in} \!= \![1, 1]$ (near the input) to $\bm{p}^{out} \!=\! [P, P]$ (near the output).

Visualization results in Fig.~{\ref{fig-vis}} reveal the distinct behaviors of \textit{t}LSTM when dealing with different tasks:
(i) Wikipedia: the input can be carried to the output location with less modification if it is sufficient to determine the next character, and vice versa;
(ii) addition: the first integer is gradually encoded into memories and then interacts (performs addition) with the second integer, producing the sum;
(iii) memorization: the network behaves like a shift register that continues to move the input symbol to the output location at the correct timestep;
(iv) sequential MNIST: the network is more sensitive to the pixel value change (representing the contour, or topology of the digit) and can gradually accumulate evidence for the final prediction;
(v) sequential \textit{p}MNIST: the network is sensitive to high value pixels (representing the foreground digit), and we conjecture that this is because the permutation destroys the topology of the digit, making each high value pixel \mbox{potentially important.}

From Fig.~{\ref{fig-vis}} we can also observe common phenomena for all tasks:
(i) at each timestep, the values at different tensor locations are markedly different, implying that wider (larger) tensors can encode more information, with less effort  to compress it;
(ii) from the input to the output, the values become increasingly distinct and are shifted by time, revealing that deep computations are indeed performed together with temporal computations, with \mbox{long-range dependencies carried by memory cells.}

\section{Related Work}

\begin{figure}[!t]
\centering
\belowcaptionskip=-4pt
%\abovecaptionskip=5pt
%\vspace{-12pt}
\includegraphics[width=0.994\textwidth]{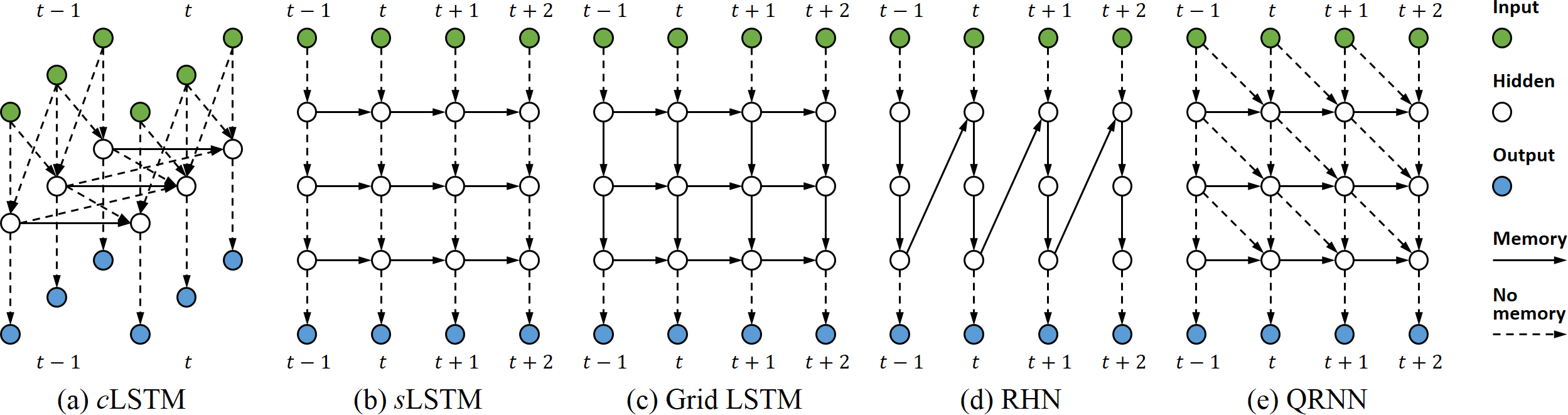}
\caption{Examples of models related to \textit{t}LSTMs.
(a) A single layer \textit{c}LSTM \cite{van2016pixel} with vector array input.
(b) A 3-layer \textit{s}LSTM \cite{graves2013generating}.
(c) A 3-layer Grid LSTM \cite{kalchbrenner2015grid}.
(d) A 3-layer RHN \cite{zilly2016recurrent}.
(e) A 3-layer QRNN \cite{bradbury2016quasi} with kernel size 2, where costly computations are done by temporal convolution.}
\label{fig-compare}
\end{figure}

\textbf{Convolutional LSTMs.}~~~
% different motivations
Convolutional LSTMs (\textit{c}LSTMs) are proposed to parallelize the computation of LSTMs when the input at each timestep is \emph{structured} (see Fig.~\ref{fig-compare}(a)), e.g., a vector array \citep{van2016pixel}, a vector matrix \citep{xingjian2015convolutional,romera2016recurrent,patraucean2015spatio,wu2016deep}, or a vector tensor \citep{stollenga2015parallel,chen2016combining}.
Unlike \textit{c}LSTMs, \textit{t}LSTM aims to increase the capacity of LSTMs when the input at each timestep is \emph{non-structured}, i.e., a single vector, and is advantageous over \textit{c}LSTMs in that:
% way of convolution
(i) it performs the convolution across different hidden layers whose structure is independent of the input structure, and integrates information bottom-up and top-down;
while \textit{c}LSTM performs the convolution within each hidden layer whose structure is coupled with the input structure, thus will fall back to the vanilla LSTM if the input at each timestep is a single vector;
% width & parameter
(ii) it can be widened efficiently without additional parameters by increasing the tensor size;
while \textit{c}LSTM can be widened by increasing the kernel size or kernel channel, which significantly increases the number of parameters;
% depth & runtime
(iii) it can be deepened with little additional runtime by delaying the output;
while \textit{c}LSTM can be deepened by using more hidden layers, which significantly increases the runtime;
% long-range dependency
(iv) it captures long-range dependencies from multiple directions through the memory cell convolution;
while \textit{c}LSTM struggles to capture long-range dependencies from multiple directions since memory cells are only gated along one direction.

\textbf{Deep LSTMs.}~~~
Deep LSTMs (\textit{d}LSTMs) extend \textit{s}LSTMs by making them deeper (see Fig.~\ref{fig-compare}(b)-(d)). To keep the parameter number small and ease training, \citet{kalchbrenner2015grid,graves2016adaptive,zilly2016recurrent,mujika2017fast} apply another RNN/LSTM along the \emph{depth} direction of \textit{d}LSTMs, which, however, multiplies the runtime.
Though there are implementations to accelerate the deep computation \citep{appleyard2016optimizing,diamos2016persistent}, they generally aim at simple architectures such \textit{s}LSTMs.
% deep & runtime
Compared with \textit{d}LSTMs, \textit{t}LSTM performs the deep computation with little additional runtime,
% feed-back
and employs a cross-layer convolution to enable the feedback mechanism.
% capacity
Moreover, the capacity of \textit{t}LSTM can be increased more efficiently by using higher-dimensional tensors, whereas in \textit{d}LSTM all hidden layers as a whole only equal to a 2D tensor (i.e., a stack of hidden vectors), \mbox{the dimensionality of which is fixed.}

\textbf{Other Parallelization Methods.}~~~
% their pros
Some methods \citep{kaiser2015neural,kaiser2016can,oord2016wavenet,bradbury2016quasi,lei2017training,chang2017dilated} parallelize the temporal computation of the sequence (e.g., use the temporal convolution, as in Fig.~\ref{fig-compare}(e)) during training, in which case full input/target sequences are accessible.
% their cons
However, during the online inference when the input presents sequentially, temporal computations can no longer be parallelized and will be blocked by deep computations of each timestep, making these methods potentially unsuitable for real-time applications that demand a high sampling/output frequency.
% our pros
Unlike these methods, \textit{t}LSTM can speed up not only training but also online inference for many tasks since it performs the deep computation by the temporal computation, which is also human-like: we convert each signal to an action and \emph{meanwhile} receive new signals in a non-blocking way.
% our cons
 Note that for the online inference of tasks that use the previous output $\bm{y}_{t-1}$ for the current input $\bm{x}_{t}$ (e.g., autoregressive sequence generation), \textit{t}LSTM cannot parallel the deep computation since it requires to delay $L\!-\!1$ timesteps to get $\bm{y}_{t-1}$.

\section{Conclusion}

We introduced the Tensorized LSTM, which employs tensors to share parameters and utilizes the temporal computation to perform the deep computation for sequential tasks.
We validated our model on a variety of tasks, showing its potential over other popular approaches.
%Our next goal is to empower \textit{t}LSTM with the ability to handle the structured input signal (e.g., the video frame) at each timestep.

\section*{Acknowledgements}

This work is supported by the NSFC grant 91220301, the Alan Turing Institute under the EPSRC grant EP/N510129/1, and the China Scholarship Council.

\bibliographystyle{named}
{\small
\bibliography{nips_2017}}

\begin{thebibliography}{}

\bibitem[\protect\citeauthoryear{Appleyard \bgroup \em et al.\egroup
  }{2016}]{appleyard2016optimizing}
Jeremy Appleyard, Tomas Kocisky, and Phil Blunsom.
\newblock Optimizing performance of recurrent neural networks on gpus.
\newblock {\em arXiv preprint arXiv:1604.01946}, 2016.

\bibitem[\protect\citeauthoryear{Arjovsky \bgroup \em et al.\egroup
  }{2016}]{arjovsky2016unitary}
Martin Arjovsky, Amar Shah, and Yoshua Bengio.
\newblock Unitary evolution recurrent neural networks.
\newblock In {\em ICML}, 2016.

\bibitem[\protect\citeauthoryear{Ba \bgroup \em et al.\egroup
  }{2016}]{ba2016layer}
Jimmy~Lei Ba, Jamie~Ryan Kiros, and Geoffrey~E Hinton.
\newblock Layer normalization.
\newblock {\em arXiv preprint arXiv:1607.06450}, 2016.

\bibitem[\protect\citeauthoryear{Bengio \bgroup \em et al.\egroup
  }{1994}]{bengio1994learning}
Yoshua Bengio, Patrice Simard, and Paolo Frasconi.
\newblock Learning long-term dependencies with gradient descent is difficult.
\newblock {\em IEEE TNN}, 5(2):157--166, 1994.

\bibitem[\protect\citeauthoryear{Bengio}{2009}]{bengio2009learning}
Yoshua Bengio.
\newblock Learning deep architectures for ai.
\newblock {\em Foundations and trends{\textregistered} in Machine Learning},
  2009.

\bibitem[\protect\citeauthoryear{Bertinetto \bgroup \em et al.\egroup
  }{2016}]{bertinetto2016learning}
Luca Bertinetto, Jo{\~a}o~F Henriques, Jack Valmadre, Philip Torr, and Andrea
  Vedaldi.
\newblock Learning feed-forward one-shot learners.
\newblock In {\em NIPS}, 2016.

\bibitem[\protect\citeauthoryear{Bradbury \bgroup \em et al.\egroup
  }{2017}]{bradbury2016quasi}
James Bradbury, Stephen Merity, Caiming Xiong, and Richard Socher.
\newblock Quasi-recurrent neural networks.
\newblock In {\em ICLR}, 2017.

\bibitem[\protect\citeauthoryear{Chang \bgroup \em et al.\egroup
  }{2017}]{chang2017dilated}
Shiyu Chang, Yang Zhang, Wei Han, Mo~Yu, Xiaoxiao Guo, Wei Tan, Xiaodong Cui,
  Michael Witbrock, Mark Hasegawa-Johnson, and Thomas Huang.
\newblock Dilated recurrent neural networks.
\newblock In {\em NIPS}, 2017.

\bibitem[\protect\citeauthoryear{Chen \bgroup \em et al.\egroup
  }{2016}]{chen2016combining}
Jianxu Chen, Lin Yang, Yizhe Zhang, Mark Alber, and Danny~Z Chen.
\newblock Combining fully convolutional and recurrent neural networks for 3d
  biomedical image segmentation.
\newblock In {\em NIPS}, 2016.

\bibitem[\protect\citeauthoryear{Chung \bgroup \em et al.\egroup
  }{2015}]{chung2015gated}
Junyoung Chung, Caglar Gulcehre, Kyunghyun Cho, and Yoshua Bengio.
\newblock Gated feedback recurrent neural networks.
\newblock In {\em ICML}, 2015.

\bibitem[\protect\citeauthoryear{Chung \bgroup \em et al.\egroup
  }{2017}]{chung2016hierarchical}
Junyoung Chung, Sungjin Ahn, and Yoshua Bengio.
\newblock Hierarchical multiscale recurrent neural networks.
\newblock In {\em ICLR}, 2017.

\bibitem[\protect\citeauthoryear{Collobert \bgroup \em et al.\egroup
  }{2011}]{collobert2011torch7}
Ronan Collobert, Koray Kavukcuoglu, and Cl{\'e}ment Farabet.
\newblock Torch7: A matlab-like environment for machine learning.
\newblock In {\em NIPS Workshop}, 2011.

\bibitem[\protect\citeauthoryear{Cooijmans \bgroup \em et al.\egroup
  }{2017}]{cooijmans2016recurrent}
Tim Cooijmans, Nicolas Ballas, C{\'e}sar Laurent, and Aaron Courville.
\newblock Recurrent batch normalization.
\newblock In {\em ICLR}, 2017.

\bibitem[\protect\citeauthoryear{De~Brabandere \bgroup \em et al.\egroup
  }{2016}]{de2016dynamic}
Bert De~Brabandere, Xu~Jia, Tinne Tuytelaars, and Luc Van~Gool.
\newblock Dynamic filter networks.
\newblock In {\em NIPS}, 2016.

\bibitem[\protect\citeauthoryear{Denil \bgroup \em et al.\egroup
  }{2013}]{denil2013predicting}
Misha Denil, Babak Shakibi, Laurent Dinh, Nando de~Freitas, et~al.
\newblock Predicting parameters in deep learning.
\newblock In {\em NIPS}, 2013.

\bibitem[\protect\citeauthoryear{Diamos \bgroup \em et al.\egroup
  }{2016}]{diamos2016persistent}
Greg Diamos, Shubho Sengupta, Bryan Catanzaro, Mike Chrzanowski, Adam Coates,
  Erich Elsen, Jesse Engel, Awni Hannun, and Sanjeev Satheesh.
\newblock Persistent rnns: Stashing recurrent weights on-chip.
\newblock In {\em ICML}, 2016.

\bibitem[\protect\citeauthoryear{Elman}{1990}]{elman1990finding}
Jeffrey~L Elman.
\newblock Finding structure in time.
\newblock {\em Cognitive science}, 14(2):179--211, 1990.

\bibitem[\protect\citeauthoryear{Garipov \bgroup \em et al.\egroup
  }{2016}]{garipov2016ultimate}
Timur Garipov, Dmitry Podoprikhin, Alexander Novikov, and Dmitry Vetrov.
\newblock Ultimate tensorization: compressing convolutional and fc layers
  alike.
\newblock In {\em NIPS Workshop}, 2016.

\bibitem[\protect\citeauthoryear{Gers \bgroup \em et al.\egroup
  }{2000}]{gers2000learning}
Felix~A Gers, J{\"u}rgen Schmidhuber, and Fred Cummins.
\newblock Learning to forget: Continual prediction with lstm.
\newblock {\em Neural computation}, 12(10):2451--2471, 2000.

\bibitem[\protect\citeauthoryear{Graves \bgroup \em et al.\egroup
  }{2013}]{graves2013speech}
Alex Graves, Abdel-rahman Mohamed, and Geoffrey Hinton.
\newblock Speech recognition with deep recurrent neural networks.
\newblock In {\em ICASSP}, 2013.

\bibitem[\protect\citeauthoryear{Graves}{2013}]{graves2013generating}
Alex Graves.
\newblock Generating sequences with recurrent neural networks.
\newblock {\em arXiv preprint arXiv:1308.0850}, 2013.

\bibitem[\protect\citeauthoryear{Graves}{2016}]{graves2016adaptive}
Alex Graves.
\newblock Adaptive computation time for recurrent neural networks.
\newblock {\em arXiv preprint arXiv:1603.08983}, 2016.

\bibitem[\protect\citeauthoryear{Ha \bgroup \em et al.\egroup
  }{2017}]{ha2016hypernetworks}
David Ha, Andrew Dai, and Quoc~V Le.
\newblock Hypernetworks.
\newblock In {\em ICLR}, 2017.

\bibitem[\protect\citeauthoryear{Hochreiter and
  Schmidhuber}{1997}]{hochreiter1997long}
Sepp Hochreiter and J{\"u}rgen Schmidhuber.
\newblock Long short-term memory.
\newblock {\em Neural computation}, 9(8):1735--1780, 1997.

\bibitem[\protect\citeauthoryear{Hutter}{2012}]{hutterhuman}
Marcus Hutter.
\newblock The human knowledge compression contest.
\newblock {\em URL http://prize.hutter1.net}, 2012.

\bibitem[\protect\citeauthoryear{Irsoy and Cardie}{2015}]{irsoy2014modeling}
Ozan Irsoy and Claire Cardie.
\newblock Modeling compositionality with multiplicative recurrent neural
  networks.
\newblock In {\em ICLR}, 2015.

\bibitem[\protect\citeauthoryear{Jozefowicz \bgroup \em et al.\egroup
  }{2015}]{jozefowicz2015empirical}
Rafal Jozefowicz, Wojciech Zaremba, and Ilya Sutskever.
\newblock An empirical exploration of recurrent network architectures.
\newblock In {\em ICML}, 2015.

\bibitem[\protect\citeauthoryear{Kaiser and Bengio}{2016}]{kaiser2016can}
{\L}ukasz Kaiser and Samy Bengio.
\newblock Can active memory replace attention?
\newblock In {\em NIPS}, 2016.

\bibitem[\protect\citeauthoryear{Kaiser and Sutskever}{2016}]{kaiser2015neural}
{\L}ukasz Kaiser and Ilya Sutskever.
\newblock Neural gpus learn algorithms.
\newblock In {\em ICLR}, 2016.

\bibitem[\protect\citeauthoryear{Kalchbrenner \bgroup \em et al.\egroup
  }{2016}]{kalchbrenner2015grid}
Nal Kalchbrenner, Ivo Danihelka, and Alex Graves.
\newblock Grid long short-term memory.
\newblock In {\em ICLR}, 2016.

\bibitem[\protect\citeauthoryear{Kingma and Ba}{2015}]{kingma2014adam}
Diederik Kingma and Jimmy Ba.
\newblock Adam: A method for stochastic optimization.
\newblock In {\em ICLR}, 2015.

\bibitem[\protect\citeauthoryear{Krause \bgroup \em et al.\egroup
  }{2017}]{krause2016multiplicative}
Ben Krause, Liang Lu, Iain Murray, and Steve Renals.
\newblock Multiplicative lstm for sequence modelling.
\newblock In {\em ICLR Workshop}, 2017.

\bibitem[\protect\citeauthoryear{Le \bgroup \em et al.\egroup
  }{2015}]{le2015simple}
Quoc~V Le, Navdeep Jaitly, and Geoffrey~E Hinton.
\newblock A simple way to initialize recurrent networks of rectified linear
  units.
\newblock {\em arXiv preprint arXiv:1504.00941}, 2015.

\bibitem[\protect\citeauthoryear{LeCun \bgroup \em et al.\egroup
  }{1989}]{lecun1989backpropagation}
Yann LeCun, Bernhard Boser, John~S Denker, Donnie Henderson, Richard~E Howard,
  Wayne Hubbard, and Lawrence~D Jackel.
\newblock Backpropagation applied to handwritten zip code recognition.
\newblock {\em Neural computation}, 1(4):541--551, 1989.

\bibitem[\protect\citeauthoryear{LeCun \bgroup \em et al.\egroup
  }{1998}]{lecun1998gradient}
Yann LeCun, L{\'e}on Bottou, Yoshua Bengio, and Patrick Haffner.
\newblock Gradient-based learning applied to document recognition.
\newblock {\em Proceedings of the IEEE}, 86(11):2278--2324, 1998.

\bibitem[\protect\citeauthoryear{Lei and Zhang}{2017}]{lei2017training}
Tao Lei and Yu~Zhang.
\newblock Training rnns as fast as cnns.
\newblock {\em arXiv preprint arXiv:1709.02755}, 2017.

\bibitem[\protect\citeauthoryear{Leifert \bgroup \em et al.\egroup
  }{2016}]{leifert2016cells}
Gundram Leifert, Tobias Strau{\ss}, Tobias Gr{\"u}ning, Welf Wustlich, and
  Roger Labahn.
\newblock Cells in multidimensional recurrent neural networks.
\newblock {\em JMLR}, 17(1):3313--3349, 2016.

\bibitem[\protect\citeauthoryear{Mujika \bgroup \em et al.\egroup
  }{2017}]{mujika2017fast}
Asier Mujika, Florian Meier, and Angelika Steger.
\newblock Fast-slow recurrent neural networks.
\newblock In {\em NIPS}, 2017.

\bibitem[\protect\citeauthoryear{Novikov \bgroup \em et al.\egroup
  }{2015}]{novikov2015tensorizing}
Alexander Novikov, Dmitrii Podoprikhin, Anton Osokin, and Dmitry~P Vetrov.
\newblock Tensorizing neural networks.
\newblock In {\em NIPS}, 2015.

\bibitem[\protect\citeauthoryear{Oord \bgroup \em et al.\egroup
  }{2016}]{oord2016wavenet}
Aaron van~den Oord, Sander Dieleman, Heiga Zen, Karen Simonyan, Oriol Vinyals,
  Alex Graves, Nal Kalchbrenner, Andrew Senior, and Koray Kavukcuoglu.
\newblock Wavenet: A generative model for raw audio.
\newblock {\em arXiv preprint arXiv:1609.03499}, 2016.

\bibitem[\protect\citeauthoryear{Patraucean \bgroup \em et al.\egroup
  }{2016}]{patraucean2015spatio}
Viorica Patraucean, Ankur Handa, and Roberto Cipolla.
\newblock Spatio-temporal video autoencoder with differentiable memory.
\newblock In {\em ICLR Workshop}, 2016.

\bibitem[\protect\citeauthoryear{Romera-Paredes and
  Torr}{2016}]{romera2016recurrent}
Bernardino Romera-Paredes and Philip Hilaire~Sean Torr.
\newblock Recurrent instance segmentation.
\newblock In {\em ECCV}, 2016.

\bibitem[\protect\citeauthoryear{Rumelhart \bgroup \em et al.\egroup
  }{1986}]{rumelhart1986learning}
David~E Rumelhart, Geoffrey~E Hinton, and Ronald~J Williams.
\newblock Learning representations by back-propagating errors.
\newblock {\em Nature}, 323(6088):533--536, 1986.

\bibitem[\protect\citeauthoryear{Schmidhuber}{1992}]{schmidhuber1992learning2}
J{\"u}rgen Schmidhuber.
\newblock Learning to control fast-weight memories: An alternative to dynamic
  recurrent networks.
\newblock {\em Neural Computation}, 4(1):131--139, 1992.

\bibitem[\protect\citeauthoryear{Stollenga \bgroup \em et al.\egroup
  }{2015}]{stollenga2015parallel}
Marijn~F Stollenga, Wonmin Byeon, Marcus Liwicki, and Juergen Schmidhuber.
\newblock Parallel multi-dimensional lstm, with application to fast biomedical
  volumetric image segmentation.
\newblock In {\em NIPS}, 2015.

\bibitem[\protect\citeauthoryear{Sutskever \bgroup \em et al.\egroup
  }{2011}]{sutskever2011generating}
Ilya Sutskever, James Martens, and Geoffrey~E Hinton.
\newblock Generating text with recurrent neural networks.
\newblock In {\em ICML}, 2011.

\bibitem[\protect\citeauthoryear{Taylor and Hinton}{2009}]{taylor2009factored}
Graham~W Taylor and Geoffrey~E Hinton.
\newblock Factored conditional restricted boltzmann machines for modeling
  motion style.
\newblock In {\em ICML}, 2009.

\bibitem[\protect\citeauthoryear{van~den Oord \bgroup \em et al.\egroup
  }{2016}]{van2016pixel}
Aaron van~den Oord, Nal Kalchbrenner, and Koray Kavukcuoglu.
\newblock Pixel recurrent neural networks.
\newblock In {\em ICML}, 2016.

\bibitem[\protect\citeauthoryear{Wisdom \bgroup \em et al.\egroup
  }{2016}]{wisdom2016full}
Scott Wisdom, Thomas Powers, John Hershey, Jonathan Le~Roux, and Les Atlas.
\newblock Full-capacity unitary recurrent neural networks.
\newblock In {\em NIPS}, 2016.

\bibitem[\protect\citeauthoryear{Wu \bgroup \em et al.\egroup
  }{2016a}]{wu2016deep}
Lin Wu, Chunhua Shen, and Anton van~den Hengel.
\newblock Deep recurrent convolutional networks for video-based person
  re-identification: An end-to-end approach.
\newblock {\em arXiv preprint arXiv:1606.01609}, 2016.

\bibitem[\protect\citeauthoryear{Wu \bgroup \em et al.\egroup
  }{2016b}]{wu2016multiplicative}
Yuhuai Wu, Saizheng Zhang, Ying Zhang, Yoshua Bengio, and Ruslan Salakhutdinov.
\newblock On multiplicative integration with recurrent neural networks.
\newblock In {\em NIPS}, 2016.

\bibitem[\protect\citeauthoryear{Xingjian \bgroup \em et al.\egroup
  }{2015}]{xingjian2015convolutional}
SHI Xingjian, Zhourong Chen, Hao Wang, Dit-Yan Yeung, Wai-kin Wong, and
  Wang-chun Woo.
\newblock Convolutional lstm network: A machine learning approach for
  precipitation nowcasting.
\newblock In {\em NIPS}, 2015.

\bibitem[\protect\citeauthoryear{Zhang \bgroup \em et al.\egroup
  }{2016}]{zhang2016architectural}
Saizheng Zhang, Yuhuai Wu, Tong Che, Zhouhan Lin, Roland Memisevic, Ruslan~R
  Salakhutdinov, and Yoshua Bengio.
\newblock Architectural complexity measures of recurrent neural networks.
\newblock In {\em NIPS}, 2016.

\bibitem[\protect\citeauthoryear{Zilly \bgroup \em et al.\egroup
  }{2017}]{zilly2016recurrent}
Julian~Georg Zilly, Rupesh~Kumar Srivastava, Jan Koutn{\'\i}k, and J{\"u}rgen
  Schmidhuber.
\newblock Recurrent highway networks.
\newblock In {\em ICML}, 2017.

\end{thebibliography}
%\bibliography{nips_2017}

\appendix

\section{Mathematical Definition for Cross-Layer Convolutions}

\subsection{Hidden State Convolution}\label{app_convh}
The hidden state convolution in (\ref{eq_trnn_conv}) is defined as:
\begin{equation}\label{hconv}
A_{t, p, m^o}  =  \sum_{k = 1}^K \lt(\sum_{m^i = 1}^{M^i}  H^{cat}_{t-1, p - \frac{K - 1}{2} + k, m^i} \cdot W_{k, m^i, m^o}^h \rt)  + b^h_{m^o}
\end{equation}
where $m^o \!\in \!\{1, 2, \cdots, M^o\}$ and zero padding is applied to keep the tensor size.

\subsection{Memory Cell Convolution}\label{app_convc}
The memory cell convolution in (\ref{eq_tlstm_mem_conv}) is defined as:
\begin{equation}\label{cconv}
 C^{conv}_{t-1, p, m} = \sum_{k = 1}^{K} C_{t-1, p - \frac{K - 1}{2} + k, m} \cdot W^c_{t, k, 1, 1}(p)
\end{equation}
To prevent the stored information from being flushed away, $\bm{C}_{t-1}$ is padded with the replication of its boundary values instead of zeros or input projections.

\section{Derivation for the Constraint of $L$, $P$, and $K$}\label{app_const}

\begin{figure}[!h]
\centering
%\belowcaptionskip=0pt
%\abovecaptionskip=9pt
%\vspace{-10pt}
\includegraphics[width=0.47\textwidth]{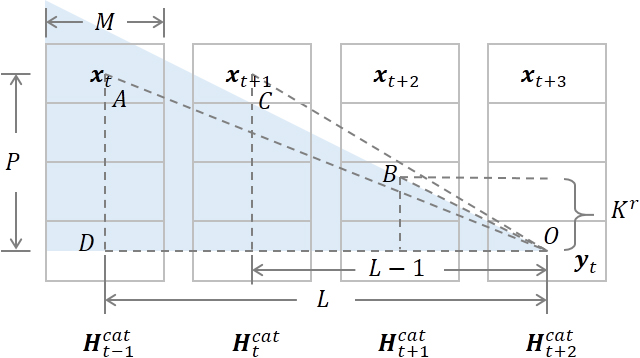}
\caption{Illustration of calculating the constraint of $L$, $P$, and $K$.
Each column is a concatenated hidden state tensor with tensor size $P\!+\!1\! =\!4$ and channel size $M$.
The volume of the output receptive field (blue region) is determined by the kernel radius $K^r$.
The output $\bm{y}_t$ for current timestep $t$ is delayed by $L\!-\!1\!=\!2$ timesteps.}
\label{fig-delay}
\end{figure}

Here we derive the constraint of $L$, $P$, and $K$ that is defined in (\ref{eq_delay}).
The kernel center location is ceiled in case that the kernel size $K$ is not odd.
Then, the kernel radius $K^r$ can be calculated by:
\begin{align}\label{eq_kr}
K^r = \frac{K - K \bmod 2}{2}
\end{align}
As shown in Fig.~\ref{fig-delay}, to guarantee the receptive field of $\bm{y}_t$ covers $\bm{x}_{1:t}$ while does not cover $\bm{x}_{t+1:T}$, the following constraint should be satisfied:
\begin{align}\label{eq_ang}
\tan \angle \textsc{aod} \leqslant \tan \angle \textsc{bod} < \tan \angle \textsc{cod}
\end{align}
which means:
\begin{align}\label{eq_kr_const}
\frac{P}{L} \leqslant \frac{K^r}{1} < \frac{P}{L - 1}
\end{align}
Plugging (\ref{eq_kr}) into (\ref{eq_kr_const}), we get:
\begin{align}\label{eq_const}
 L = \ceil[\Big]{\frac{2 P}{K - K \bmod 2}}
\end{align}

\section{Memory Cell Convolution Helps to Prevent the Vanishing/Exploding Gradients}\label{app_grad}

\citet{leifert2016cells} have proved that the \emph{lambda gate}, which is very similar to our memory cell convolution kernel, can help to prevent the vanishing/exploding gradients (see Theorem 17-18 in \citep{leifert2016cells}).
The differences between our approach and their \emph{lambda gate} are:
(i) we normalize the kernel values though a softmax function, while they normalize the gate values by dividing them with their sum, and
(ii) we share the kernel for all channels, while they do not.
However, as neither modifications affects the conditions of validity for Theorem 17-18 in \citep{leifert2016cells}, our memory cell convolution can also help to prevent the vanishing/exploding gradients.

\section{Training Details}\label{app_train}

\subsection{Objective Function}
The training objective is to minimize the negative log-likelihood (NLL) of the training sequences w.r.t. the parameter $\bm{\theta}$ (vectorized), i.e.,
\begin{equation}\label{eq_loss}
\min_{\bm{\theta}}  \frac{1}{N} \sum_{n=1}^N \sum_{t=1}^{T_n} -\ln p(\bm{y}_{n,t}^{d} | f(\bm{x}_{n,1:t}^{d}; \bm{\theta} ))
\end{equation}
where $N$ is the number of training sequences, $T_n$ the length of the $n$-th training sequence, and $p(\bm{y}_{n,t}^{d} | f(\bm{x}_{n,1:t}^{d}; \bm{\theta})) $ the likelihood of target $\bm{y}_{n,t}^{d}$ conditioned on its prediction $\bm{y}_{n,t} = f(\bm{x}_{n,1:t}^{d}; \bm{\theta})$.
Since all experiment are classification problems, $\bm{y}_{n,t}^d$ is represented as the one-hot encoding of the class label,
and the output function $\varphi(\cdot)$ is defined as a softmax function, which is used to generate the class distribution $\bm{y}_{n,t}$.
Then, the likelihood can be calculated by $p(\bm{y}_{n,t}^d |\bm{y}_{n,t}) = {y}_{n,t,s}|_{y_{n,t,s}^d = 1} $.

\subsection{Common Settings}
% common setting
In all tasks, the NLL (see (\ref{eq_loss})) is used as the training objective and is minimized by Adam \citep{kingma2014adam} with a learning rate of 0.001.
Forget gate biases are set to 4 for image classification tasks and 1  \citep{jozefowicz2015empirical} for others.
 All models are implemented by Torch7 \citep{collobert2011torch7} and accelerated by cuDNN on Tesla K80 GPUs.

We only apply CN to the output of the \textit{t}LSTM hidden state as we have tried different combinations and found this is the most robust way that can always improve the performance for all tasks.
With CN, the output of hidden state becomes:
\begin{equation}\label{cno}
\bm{H}_t = \phi \lt(\op{CN}\lt(\bm{C}_t; \bm{\Gamma}, \bm{B}\rt)\rt) \odot \bm{O}
\end{equation}

\subsection{Wikipedia Language Modeling}

As in \citep{chung2015gated}, we split the dataset into 90M/5M/5M for training/validation/test.
In each iteration, we feed the model with a mini-batch of 100 subsequences of length 50.
During the forward pass, the hidden values at the last timestep are preserved to initialize the next iteration.
We terminate training after 50 epochs.

\subsection{Algorithmic Tasks}

Following \citep{kalchbrenner2015grid}, for both tasks we randomly generate 5M samples for training and 100 samples for test, and set the mini-batch size to 15.
Training proceeds for at most 1 epoch\footnote{To simulate the online learning process, we use all training samples only once.} and will be terminated if $100\%$ test accuracy is achieved.

\subsection{MNIST Image Classification}
We set the mini-batch size to 50 and use early stopping for training.
The training loss is calculated at the last timestep.

\end{document}